\documentclass[sigconf]{acmart}

\AtBeginDocument{%
  \providecommand\BibTeX{{%
    \normalfont B\kern-0.5em{\scshape i\kern-0.25em b}\kern-0.8em\TeX}}}

\copyrightyear{2020}
\acmYear{2020}
\setcopyright{acmlicensed}
\acmConference[LAK '20]{Proceedings of the 10th International Conference on Learning Analytics and Knowledge}{March 23--27, 2020}{Frankfurt, Germany}
\acmBooktitle{Proceedings of the 10th International Conference on Learning Analytics and Knowledge (LAK '20), March 23--27, 2020, Frankfurt, Germany}
\acmPrice{15.00}
\acmDOI{10.1145/3375462.3375517}
\acmISBN{978-1-4503-7712-6/20/03}

\usepackage{subcaption}

\newcommand{\eg}{e.g.,}
\newcommand{\ie}{i.e.,}
\newcommand{\ca}{\textit{Cloud Academy}}
\newcommand{\linkcode}{\url{https://github.com/lucabenedetto/r2de-nlp-to-estimating-irt-parameters}}
\newcommand{\wt}{$N_W$}
\newcommand{\modelname}{R2DE}
\newcommand{\encitem}{\textit{question\_only}}
\newcommand{\enccorrect}{\textit{question\_correct}}
\newcommand{\encfull}{\textit{question\_full}}

\newcommand{\namepoli}{Politecnico di Milano}
\newcommand{\nameca}{Cloud Academy Sagl}

\begin{document}

\title{\modelname{}: a NLP approach to estimating IRT parameters of newly generated questions}

\author{Luca Benedetto}
\email{luca.benedetto@polimi.it}
\orcid{0000-0002-5113-4696}
\affiliation{%
  \institution{\namepoli{}}
  \city{Milan}
  \state{Italy}
}

\author{Andrea Cappelli}
\email{andrea.cappelli@cloudacademy.com}
\affiliation{%
  \institution{\nameca{}}
  \city{Mendrisio}
  \state{Switzerland}
}

\author{Roberto Turrin}
\email{roberto.turrin@cloudacademy.com}
\affiliation{%
  \institution{\nameca{}}
  \city{Mendrisio}
  \state{Switzerland}
}

\author{Paolo Cremonesi}
\email{paolo.cremonesi@polimi.it}
\orcid{0000-0002-1253-8081}
\affiliation{%
  \institution{\namepoli{}}
  \city{Milan}
  \state{Italy}
}

\renewcommand{\shortauthors}{Benedetto, et al.}

\begin{abstract}
The main objective of exams consists in performing an assessment of students' expertise on a specific subject.
Such expertise, also referred to as skill or knowledge level, can then be leveraged in different ways (\eg{} to assign a grade to the students, to understand whether a student might need some support, etc.).
Similarly, the questions appearing in the exams have to be assessed in some way before being used to evaluate students.
Standard approaches to questions' assessment are either subjective (\eg{} assessment by human experts) or introduce a long delay in the process of question generation (\eg{} pretesting with real students).
In this work we introduce \modelname{} (which is a \textit{Regressor for Difficulty and Discrimination Estimation}), a model capable of assessing newly generated multiple-choice questions by looking at the text of the question and the text of the possible choices.
In particular, it can estimate the difficulty and the discrimination of each question, as they are defined in Item Response Theory.
We also present the results of extensive experiments we carried out on a real world large scale dataset coming from an e-learning platform, showing that our model can be used to perform an initial assessment of newly created questions and ease some of the problems that arise in question generation.
\end{abstract}

\ccsdesc[500]{Computing methodologies~Natural language processing}
\ccsdesc[500]{Computing methodologies~Ensemble methods}
\ccsdesc[300]{Computing methodologies~Information extraction}
\ccsdesc[500]{Applied computing~Education}

\keywords{learning analytics, natural language processing, knowledge tracing, item response theory, latent traits estimation, educational data mining}

\maketitle

\section{Introduction}\label{intro}
Being able to estimate with a low degree of uncertainty the knowledge level of a student is crucial for providing effective material - tailored to his expertise - and thus improving the learning experience.
The task of estimating the skill level of a student by analysing the results of his interactions with assessment items (\ie{} questions) is known as Knowledge Tracing (KT), and it is most commonly addressed with logistic models or neural networks \cite{abyaa2019learner}.

Although Deep Knowledge Tracing (DKT) \cite{piech2015deep} - which is KT performed by means of neural networks - generally provides the highest accuracy in predicting the results of future answers \cite{zhang2017dynamic,wang2019deep}, logistic models and in particular dynamic Item Response Theory (IRT) models \cite{wang2013bayesian} are still used because of their explainability.
Indeed, while in DKT the model is considered as a black-box, logistic models estimate latent traits of students and items, which enable a straightforward interpretation.
In particular, with IRT models it is possible to estimate the skill level of each student and its evolution over time, as well as the difficulty and the discrimination of each question.
The concept of difficulty is straightforward: if a question is more difficult than another, it requires a higher skill level to be answered correctly with the same probability.
On the other hand, the discrimination determines how rapidly the odds of a correct answer increase or decrease with the skill level of the student.
Therefore, discrimination can be used as a measure of the quality of an item.
Indeed, questions with low discrimination provide little or no information about the skill level of the student answering, regardless of the difficulty of the item, because students of all skill levels have similar probability of answering correctly.
Because of this, an estimation of the discrimination offers immediate feedback to content creators, who can modify the questions accordingly, in order to produce better (\ie{} more discriminative) assessment items.

Having access to a history of exam results, the latent traits of the students and the questions involved in the exams can be estimated via likelihood maximization.
In a similar way, when a student takes an exam composed of calibrated questions (\ie{} items whose latent traits are known), it is possible to estimate the skill level of the student from the results of the exam.
Therefore, when a new question is created, it cannot be used for assessing students until a reliable estimation of its latent traits is performed.
Also, some items might prove unsuited for assessing students (\eg{} because of a discrimination which is too low or a difficulty which is too high or too low), thus having to be removed from the set of possible questions; it is important to do this as soon as possible.

A standard solution to the lack of an estimation of the latent traits of newly-created items, which is often referred to as the cold-start problem, consists in pretesting \cite{yaneva2019predicting}: before using a newly developed item for assessment, it is administered to a certain number of students (usually few hundreds of few thousands) as if it was a regular exam question, but it is not used for scoring.
On the contrary, the other questions of the exam are used for assessing the students.
Then, the estimated skill level of such students is used together with the answers given to the item under pretesting to estimate its latent traits.
Although this procedure leads indeed to an estimation of the latent traits of each item, it causes a long delay between the creation of an item and being able to use it for assessing students, and it also increases the development costs.

Another solution to the cold-start problem consists in using latent traits manually set by human experts: this approach enables the immediate usage of newly created questions in tests for assessing students, but it introduces a high uncertainty in the estimation, due to its nature intrinsically subjective.

In this work we introduce \modelname{} (a \textit{Regressor for Difficulty and Discrimination Estimation}), a model that is capable of estimating the difficulty and the discrimination (as defined by Item Response Theory) of multiple-choice questions from the text of the questions and the text of the possible options.
We present the results of extensive experiments performed on a real-world large scale dataset coming from an e-learning platform, showing that this model leads to a good estimation of the latent traits of new items, reducing both the importance of pretesting (making it necessary only for a fine-tuning of the initial estimation) and the uncertainty of the estimation of the latent traits (in comparison with latent traits manually set by human experts).
Thus, it can be used as a first assessment of the difficulty and the discrimination of newly created questions, enabling an immediate usage in assessment tests and reducing the number of questions that have to be dropped after pretesting because of quality issues.

The contributions of this work are:
i) the introduction of a novel model which uses natural language for estimating IRT latent traits of assessment items;
ii) extensive testing of the model on a real world large scale dataset coming from an e-learning platform;
iii) publication of the code used for implementing and testing this model, at \linkcode{}.

The rest of the document is organized as follows: after an introduction to the current state of the art and the related works in Section \ref{related_work}, Section \ref{background} focuses on IRT and on the performance prediction task in order to establish a common ground.
\modelname{} is then presented in Section \ref{model_sec}, followed by an introduction to the experimental dataset in Section \ref{dataset} and a preliminary experiment for model choice and hyperparameter tuning in Section \ref{model_choice}.
The results of the experiments are shown is Section \ref{results} and, lastly, Section \ref{conclusions} concludes the paper.

\section{Related Work}\label{related_work}
The related work can be classified in the following categories: i) research about Knowledge Tracing (KT) and ii) research about Natural Language Processing (NLP) approaches for the estimation of questions' latent traits.

\subsection{Knowledge Tracing}
The concept of Knowledge Tracing (KT) was pioneered many years ago by Atkinson \cite{atkinson1972ingredients}, but it is still extensively explored in research.
As reported in a recent review by Abyaa et al. \cite{abyaa2019learner}, the methods that are most commonly used in KT are logistic models and neural networks.
Belonging to the family of logistic models are the approaches based on Item Response Theory (IRT) (\eg{} dynamic IRT \cite{wang2013bayesian}) and the Elo rating system \cite{verhagen2019toward}.

Recent literature claims that Deep Knowledge Tracing (DKT) - which was first introduced by Piech et al. in \cite{piech2015deep} and consists in performing KT by means of neural networks - outperforms logistic models in predicting the results of future exams \cite{zhang2017dynamic,chen2018prerequisite,abdelrahman2019knowledge,zhang2017incorporating}, but this advantage is not agreed across the community \cite{mao2018deep,dingdeep,yeung2018addressing,wilson2016back}.
Also, DKT does not estimate explicitly the skill level of students nor the latent traits of questions, which makes the interpretation of such models a strenuous task.
There have been some attempts to make DKT explainable \cite{lee2019knowledge,yeung2019deep}, but they did not reach the same level of explainability as logistic models and are much more computationally expensive.
Therefore, because of DKT being hard to explain and more complex from a computational point of view, logistic models and in particular models based on IRT are still widely used in the literature \cite{wilson2016back,dingdeep}.

Item Response Theory \cite{hambleton1991fundamentals} estimates latent traits of students and items (\ie{} questions) involved in an exam: the simplest model, named ``Rasch model'' \cite{rasch1960studies}, associates a skill level to each student and a difficulty level to each question.
More complex models take into consideration additional latent traits \cite{loken2010estimation} (\eg{} the probability of correct answer by guessing): in this work we consider the two-parameter model, which associates a discrimination to each item.
The discrimination determines how rapidly the odds of a correct answer increase or decrease with the skill level of the student and is a measure of how well an item can discriminate between students whose skill levels are above or below a certain threshold.
Given a list of interactions between a set of students and a set of questions, latent factors of both students and questions can be estimated maximizing the likelihood of the observed results \cite{wang2010brief}.
Then, the calibrated items can be used for assessing new students.
Given a set of calibrated questions (\ie{} questions whose latent traits are known) it is possible to estimate the skill level of the students answering those questions by likelihood maximization.
Similarly, it is possible to leverage the answers of students that have already been assessed to estimate the latent traits of newly-created questions.
The cold-start problem arises as soon as a new question is generated: since the latent factors of the item are unknown and there is no history of interactions to estimate them, it cannot be used for scoring students.
A standard solution to this problem is pretesting: the newly-generated question is given to some students without being used for assessing them and, looking at the answers provided by these students and at their skill level (assessed using other questions), the latent factors of the new question can be estimated \cite{yaneva2019predicting}.
This procedure introduces a long delay between the time when a question is generated and when it can be used for assessment.
A possible solution to this problem consists in using the text of the newly-generated question to estimate its latent factors and remove pretesting from the pipeline (or, at least, reduce the number of students required for pretesting).
To the best of our knowledge, in previous works, only Huang et al. in \cite{huang2019ekt} explicitly mentioned that their method for estimating the difficulty of questions from their text could also be useful for targeting the cold-start problem.
However, their model differs from the one presented in this paper because it estimates only the difficulty, without focusing on the discrimination; also, no code is available.

\subsection{NLP for Latent Traits Estimation}
The idea of using the text of a question to predict its difficulty is not new; however, most of previous works focused on readability estimation \cite{dubay2004principles,kintsch2014reading,yaneva2019predicting}, which is a concept different from the difficulty defined in IRT.
Wang et. al in \cite{wang2014regularized} used textual information together with the interaction with users to estimate the difficulty of questions, but the focus was on Community Question Answering (CQA) services, thus considering a concept of difficulty different from the IRT-estimated difficulty used in this paper.

Closer to our paper are some more recent contributions that use NLP techniques to estimate the difficulty of assessment items, but they all define the difficulty as the fraction of wrong answers given to a question (referred to as ``wrongness'', from now on), which is less accurate than the IRT-estimated difficulty.
One of such works is \cite{huang2017question}, in which the authors propose a neural model to predict the difficulty of reading problems in Standard Tests (\ie{} problems whose answer can be found in a text that is given together with the question to the students) given the text of the document, the text of the question and the text of the possible answers.
In \cite{huang2019ekt} the authors use a neural network model to extract the Knowledge Components (\ie{} the skills) related to each question from its text.
A similar approach is adopted by Su et al. in \cite{su2018exercise}, in which the accuracy of difficulty estimation from question's text is measured by looking at the precision in predicting the performance of the students in future exams.
The main differences between these works and the present paper can be outlined in the following points: i) we use the IRT estimated difficulty as ground truth, which is more accurate than the ``wrongness as difficulty'' approach; ii) we estimate both the difficulty and the discrimination of the questions, thus providing a mean to assess the quality of newly-created questions.
\section{Theoretical Background}\label{background}
The objective of the model introduced in this work consists in using an NLP approach to estimate the difficulty and the discrimination of assessment items, as they are defined in IRT.
Such estimations are then evaluated considering i) the accuracy with respect to ground truth values of the latent traits and ii) the accuracy in the performance prediction task.
In order to establish common ground this section presents an introduction i) to IRT, providing the mathematical explanation of the concepts of difficulty and discrimination, and ii) to the performance prediction task.

\subsection{Item Response Theory}\label{background_irt}
We use a two-parameter IRT model \cite{hambleton1991fundamentals}, which is characterized by three latent traits (the ``two'' refers to the number of items' latent traits): i) a skill level $\theta$ associated to each student, ii) a difficulty level $b$, and iii) a discrimination $a$ associated to each assessment item.
These latent traits are then used to compute the probability that student $i$ correctly answers question $j$ with the item response function:
\begin{equation*}
\text{P}_{\text{CORRECT}} = \frac{1}{1 + e^{-a_j \cdot (\theta_i - b_j)}}
\end{equation*}

\begin{figure}
\begin{subfigure}[t]{\linewidth}
  \centering
  \includegraphics[width=\linewidth]{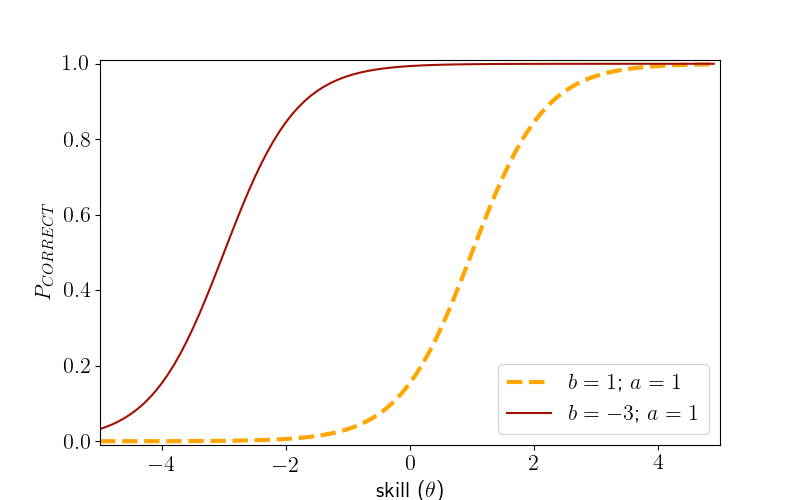}
  \caption{Same discrimination $a$, different difficulty $b$.}
  \label{fig:example_irf_diff_discr_a}
\end{subfigure}
\begin{subfigure}[b]{\linewidth}
  \centering
  \includegraphics[width=\linewidth]{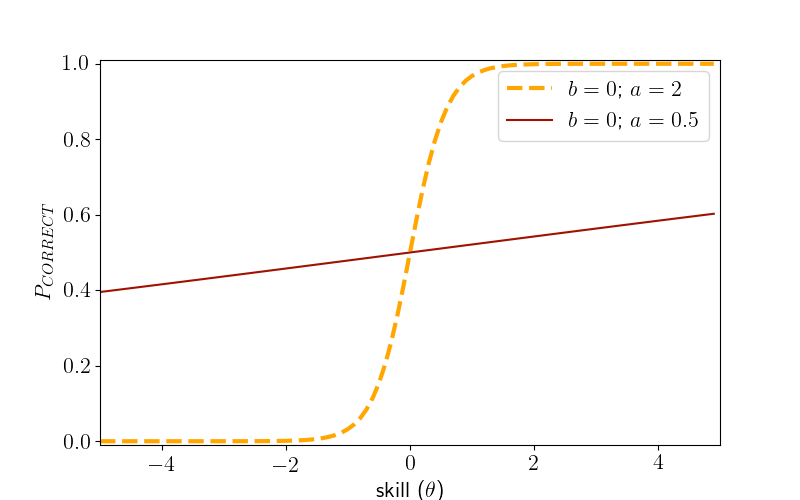}
  \caption{Same difficulty $b$, different discrimination $a$.}
  \label{fig:example_irf_diff_discr_b}
\end{subfigure}
\caption{Example showing the effects of different difficulties and discrimination on the item response functions.}
\label{fig:example_irf_diff_discr}
\Description[i.r.f. for items with different difficulty and discrimination]{item response functions of items with different difficulty and discrimination}
\end{figure}

An example of the item response functions of two questions with equal discrimination and different difficulties is displayed in Figure \ref{fig:example_irf_diff_discr_a}.
According to the intuition, students with the same skill level (represented on the x-axis) have a lower probability of answering correctly the question with higher difficulty.
The discrimination $a$, on the other hand, affects the steepness of the logistic curve, and that is the reason why it can be used as a measure of how well an item can discriminate between students whose skill level is above or below a certain threshold (\ie{} the difficulty of the question).
Figure \ref{fig:example_irf_diff_discr_b} shows the item response function of two questions with equal difficulty and different discrimination: the plot for the question with low discrimination is almost flat, showing that students with very different skill levels have similar probabilities of correctly answering the question.
Thus, the information that can be gathered from that item is very limited.

Given the correctness of the answers that a student gave to a set of calibrated assessment items $Q = \{q_1, q_2,  ..., q_{N_q}\}$, it is possible to estimate the knowledge level $\tilde{\theta}$ of the student.
This is done by maximizing the result of the multiplication between the item response functions of the questions that were answered correctly and the inverse of the item response functions of the questions that were answered erroneously, with the following formula:
\begin{equation*}
\tilde{\theta} = \max_{\theta} \left[ \prod_{q_j \in Q_{C}}\frac{1}{1 + e^{-a_j \cdot (\theta - b_j)}} \cdot \prod_{q_j \in Q_{W}} \left( 1 - \frac{1}{1 + e^{-a_j \cdot (\theta - b_j)}}\right) \right]
\end{equation*}
In the equation above $Q_{C}$ is the set of questions correctly answered and $Q_{W}$ is the set of questions that were answered erroneously.

\subsection{The Performance Prediction Task}\label{background_perf_pred}
The latent traits estimated with IRT are non-observable by definition.
Therefore, even though they can be considered as ground truth and are commonly used to evaluate the accuracy of a model (\eg{} \cite{yaneva2019predicting}), they have to be carefully dealt with.
For this reason, in this work we validate our model not only by measuring the accuracy in predicting the latent traits of assessment items, but also by measuring its effects on the performance prediction task, which is the only way to work with an observable ground truth: the correctness of students' answers.

The performance prediction task consists in predicting the correctness of the answers given by a student to a sequence of assessment items.
It can be used to measure how well the latent traits of the items are estimated by our model since these latent traits are a key element in predicting the correctness of students' answers.
Given the ordered sequence of questions that a student interacted with, the correctness of each answer and an estimation of the latent traits of the items, it is performed as follows: i) given the latent traits of the first item and the estimated skill level of the student at that time (possibly unknown, in case of the first item), the correctness of his answer is predicted; ii) the actual answer is observed and compared to the predicted value in order to measure the accuracy of the prediction; iii) the actual answer is used to update the estimation of the skill level of the student; iv) this sequence of steps is repeated for all the assessment items the student interacted with.

In this work, the performance prediction task is used to evaluate the accuracy of our model in estimating the latent traits of assessment items.
This is done by comparing the accuracy obtained in predicting the correctness of the answer using different algorithms to estimate the latent traits of the questions.

\section{\modelname{}}\label{model_sec}
\begin{figure}[h]
\centering
\includegraphics[width=75mm]{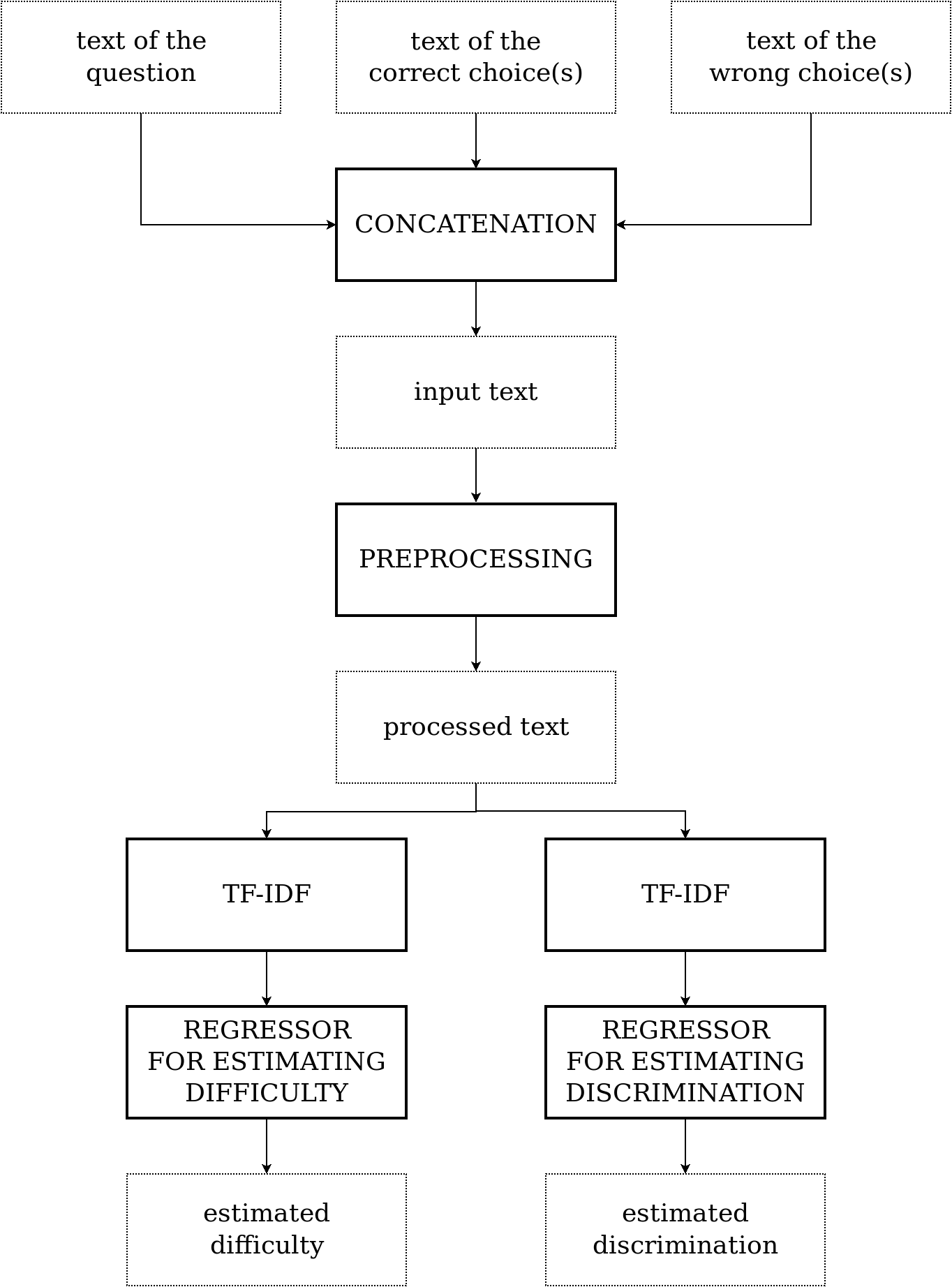}
\caption{Structure of \modelname{}, from the input question to the estimated latent traits.}\label{fig:model_structure}
\Description[Structure of R2DE]{Structure of R2DE}
\end{figure}

This section introduces \modelname{}, which is a \textit{Regressor for Difficulty and Discrimination Estimation}.
The structure of \modelname{}, from the input question to the estimated latent traits, is shown in Figure \ref{fig:model_structure}.
This section will focus on presenting the building blocks of the model and the steps that lead to the estimation of difficulty and discrimination from text.

\subsection{Data Model}\label{data_format}
The model introduced in this paper requires two different types of information.
First of all, since \modelname{} works on text of multiple-choice questions, it needs the text of all the questions and the text of all the possible choices, as well as an indication of which choice contains the correct answer to each question.
The model can also deal with the scenario where a question has multiple correct choices.
This text information is used to generate the feature arrays that are used as input to the model.

Secondly, it requires the history of interactions between a set of students and a set of assessment items.
For each interaction (\ie{} the answer given by a student to a question), four fields of information are needed: i) a unique identifier of the student, ii) a unique identifier of the assessment item, iii) the correctness of the answer, and iv) the timestamp of the interaction.
This data is used both to perform the IRT estimation of the latent traits of each question, which are used as ground truth for training \modelname{}, and to perform validation with the performance prediction task.
\subsection{Features Engineering and Target Labels}\label{data_enc}
This subsection shows the steps necessary to obtain the target labels and the feature arrays from data structured as presented in Subsection \ref{data_format}.

\subsubsection{Input Features}
The first step consists in creating an input text for each question.
In this work three approaches (later referred to as encodings) were tested: 
\begin{itemize}
\item \encitem{}: considering only the text of the questions;
\item \enccorrect{}: concatenating the text of the correct options (possibly more than one option is correct) to the text of the question;
\item \encfull{}: concatenating the text of all the possible options (both correct and wrong) to the text of the question.
\end{itemize}
Considering a fictitious example, let us assume that the student is shown the question ``\textit{Which is the capital city of Germany?}'' and the possible answers are ``\textit{London}'', ``\textit{Berlin}'', ``\textit{Madrid}'' and ``\textit{Paris}''.
Then, the body of text to represent the question would become, in the three cases: i) ``\textit{Which is the capital city of Germany}'', ii) ``\textit{Which is the capital city of Germany Berlin}'', and iii) ``\textit{Which is the capital city of Germany London Berlin Madrid Paris}''.
The outcome of this first step consists - in all three cases - in an input text characterizing each question.

The second step consists in preprocessing the corpus made of all the input texts using standard steps of NLP: removal of stop words, removal of punctuation, and stemming \cite{manning1999foundations}.

The third step consists in creating arrays of features from the input text of each item using a technique from Information Retrieval: TF-IDF (\ie{Term Frequency-Inverse Document Frequency}) \cite{manning2010introduction}.
The TF-IDF weight represents how important is a word (or a set of words) to a document in a corpus.
The importance grows with the number of occurrences of the word in the document but it is limited by its frequency in the whole corpus: intuitively, words that are very frequent in all the documents of the corpus are not important to any of them.
In particular, the formula used to compute the TF-IDF weight of word $w$ in document $d$ belonging to the corpus $C = \{d_1, d_2, ..., d_{N_d}\}$ is the following
\begin{equation*}
\text{TFIDF}(w, d, C) = \text{count}(w, d) \cdot \left( \log_e \frac{N_d + 1}{\text{count}(w,C)+1} + 1\right)
\end{equation*}
where $\text{count}(w, d)$ is the number of occurrence of $w$ in document $d$, $N_d$ is the number of documents in the corpus $C$ and $\text{count}(w, C)$ is the number of documents in the corpus $C$ where $w$ appears.

Lastly, since the feature set produced as outcome of TF-IDF is too large to be directly used as input of \modelname{} (\ie{} one feature for each stemmed word generated from the original corpus) and standard approaches for dimensionality reduction such as Principal Component Analysis (PCA) would heavily affect the possibility of understanding the impact of specific concepts on the latent traits of the questions, a simpler method was used to reduce the size of the feature set.
Specifically, only the top \wt{} features are considered: this is done by sorting the features (\ie{} the stemmed words as obtained with preprocessing) according to their number of occurrences in the corpus and keeping only the \wt{} most frequent ones.
This threshold (\wt{}) is considered as one of the parameters of the model and therefore can be chosen with cross-validation in order to be tuned for a specific dataset (as will be shown in Section \ref{exp_setup}).

\subsubsection{Target Labels}
The target labels are the latent traits (specifically, difficulty and discrimination as defined in IRT) of the items in the dataset.
The latent traits can be estimated from the history of interactions between students and questions with a two-parameter IRT model.

\subsection{The Regression Algorithm}\label{model}
\modelname{} contains two regressors that work in parallel to estimate i) the difficulty and ii) the discrimination of multi-choice questions.
For both the elements a set of different algorithms should be tested in order to choose the ones that perform better on a specific dataset.
Specifically, we tested Random Forests, Decision Trees, Support Vector Regression, and Linear Regression.

Using the same approach as in \cite{yaneva2019predicting}, model choice and tuning of the parameters are performed with 5-fold cross validation.
Differently from previous works, we also use cross validation to choose which one of the three encodings described in Subsection \ref{data_enc} - \ie{} i) question's text only, ii) question's text and correct answer's text, iii) question's text plus text of all the possible answers - to use and \wt{}, the number of most frequent keywords that should be used for the estimation.
The fact that we can tune all these parameters makes this model more flexible and likely to perform comparably well on several datasets.

\section{Experimental Dataset}\label{dataset}
To the best of our knowledge there are no public datasets containing both the text of the questions and the results of the answers, and all previous works experimented on private data collections.
Our model as well is evaluated on a private database, which is a sample of actual data provided by the e-learning provider \ca{}\footnote{\url{https://cloudacademy.com/}}.
In particular, two data collections are used in this work, according to the required data format described in Subsection \ref{data_format}: i) one contains the information about the assessment items, ii) the other contains the information about the interactions between the students and the questions (\ie{} the answers given by the students to the assessment items).

\subsection{Question Dataset}
In total, there are about 10K questions and the average number of possible choices is 4.2; the distribution of questions per number of choices is displayed in Table \ref{table:num_answes}.
In any case, regardless of the number of possible choices, when a question is prompted to a student only four of the possible choices are shown, among which there are always the correct ones.
\begin{table}[ht]
\caption{Distribution of questions per number of possible choices.}
\centering
\begin{tabular}{c c}
\hline
\# of choices & percentage of items\\
\hline
4 & 86\%\\
5 & 8\%\\
6 & 5\%\\
> 6 & < 1\%\\
\hline
\end{tabular}
\label{table:num_answes}
\end{table}

The average length of the questions is 26.75 words and the answers are on average 6.83 words long, but the length of both the questions and the possible choices varies considerably.
Table \ref{table:len_items} presents the distribution of questions and Table \ref{table:len_choices} presents the distribution of choices per length.
\begin{table}[ht]
\caption{Distribution of questions per length.}
\centering
\begin{tabular}{c c}
\hline
length (\# of words) & percentage of items\\
\hline
len <= 5 & 1\% \\
5 < len <= 10 & 10\% \\
10 < len <= 20 & 39\% \\
20 < len <= 50 & 37\% \\
len > 50 & 13\% \\
\hline
\end{tabular}
\label{table:len_items}
\end{table}

\begin{table}[ht]
\caption{Distribution of possible choices per length.}
\centering
\begin{tabular}{c c}
\hline
length (\# of words) & percentage of choices\\
\hline
len = 1 & 21\% \\
1 < len <= 5 & 35\% \\
5 < len <= 10 & 20\% \\
len > 10 & 24\% \\
\hline
\end{tabular}
\label{table:len_choices}
\end{table}

\subsection{Interactions Dataset}
The interaction dataset used for the experiments contains about 2.3M interactions, collected over two years and involving a total of about 17K distinct students and 8K distinct assessment items.
Overall, the interactions with a correct answer are the 64.69\%, but this varies considerably depending on the specific students and items.
This dataset was built from the original sample of data provided by \ca{} in order i) to contain only ``first timers'' (\ie{} for each student-item pair, only the answer given during the first attempt is considered), and ii) to contain only questions that were answered by at least 100 distinct students.
This was done for two different reasons: i) students that answer several times a specific question are more likely to answer correctly, thus impacting the accuracy of the latent traits estimated with IRT; ii) since the objective of this work consists in using textual information for estimating the latent traits of questions and not in analyzing the effectiveness of IRT, working on items with low support would affect the IRT estimation of questions' latent traits.
Having the most possible accurate IRT estimation of difficulty and discrimination is crucial: in fact, the IRT-estimated latent traits are considered as ground truth while training \modelname{} and, in case of a bad estimation, we would train our model with noisy samples.

\begin{figure}
\begin{subfigure}[t]{\linewidth}
  \centering
  \includegraphics[width=\linewidth]{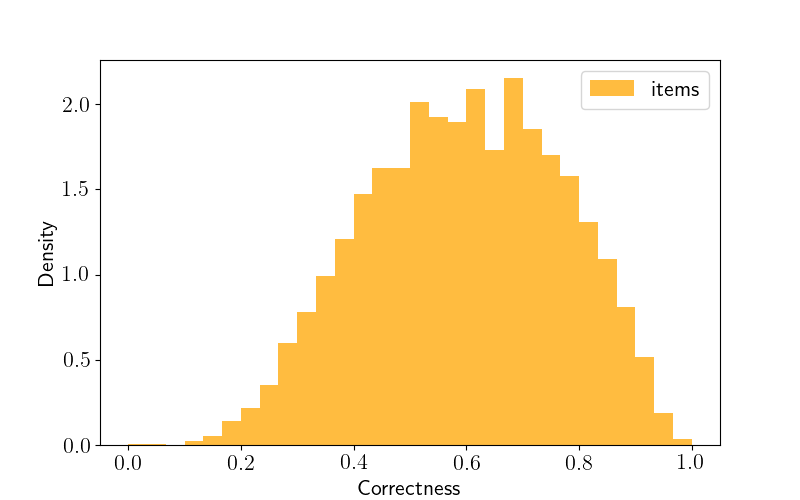}
  \caption{Distribution of questions per correctness.}
  \label{fig:distribution_items_after}
\end{subfigure}
\begin{subfigure}[b]{\linewidth}
  \centering
  \includegraphics[width=\linewidth]{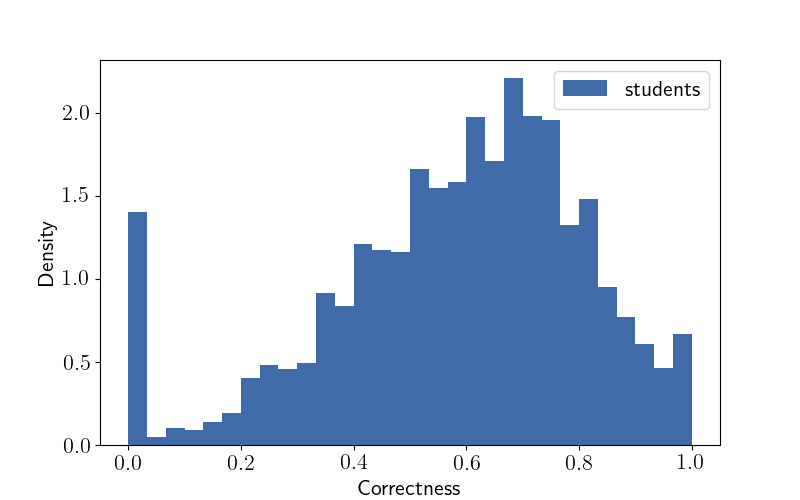}
  \caption{Distribution of students per correctness.}
  \label{fig:distribution_students_after}
\end{subfigure}
\caption{Distribution of students and items per correctness after filtering questions answered by less than 100 students.}
\label{fig:distribution}
\Description[Distribution of students and items per correctness]{distribution of students and items per correctness}
\end{figure}

Figure \ref{fig:distribution} displays the distribution of students and questions per correctness, showing that both present a Gaussian-shaped distribution, although - in the case of students - the distribution shows two peaks for values of correctness close to 0 and 1.
This is probably due to the fact that some students have low support (\ie{} they are involved in few interactions).
However, differently from what was done for the items, students with low support were not removed in order not to reduce too much the size of the dataset.

Some additional statistics about the dataset are presented in Table \ref{table:dataset}.

\begin{table}[ht]
\caption{Statistics about the \ca{} dataset, after data cleaning.}
\centering
\begin{tabular}{c c c}
\hline
value & mean & std. dev.\\
\hline
\# interactions per student & 130.54 & 193.19 \\
\# interactions per item & 283.03 & 407.47 \\
correctness per student & 58.25\% & 22.86\% \\
correctness per item & 59.46\% & 17.30\% \\
\hline
\end{tabular}
\label{table:dataset}
\end{table}

\section{Experimental Setup}\label{exp_setup}
\begin{figure*}[ht]
\centering
\includegraphics[width=\textwidth]{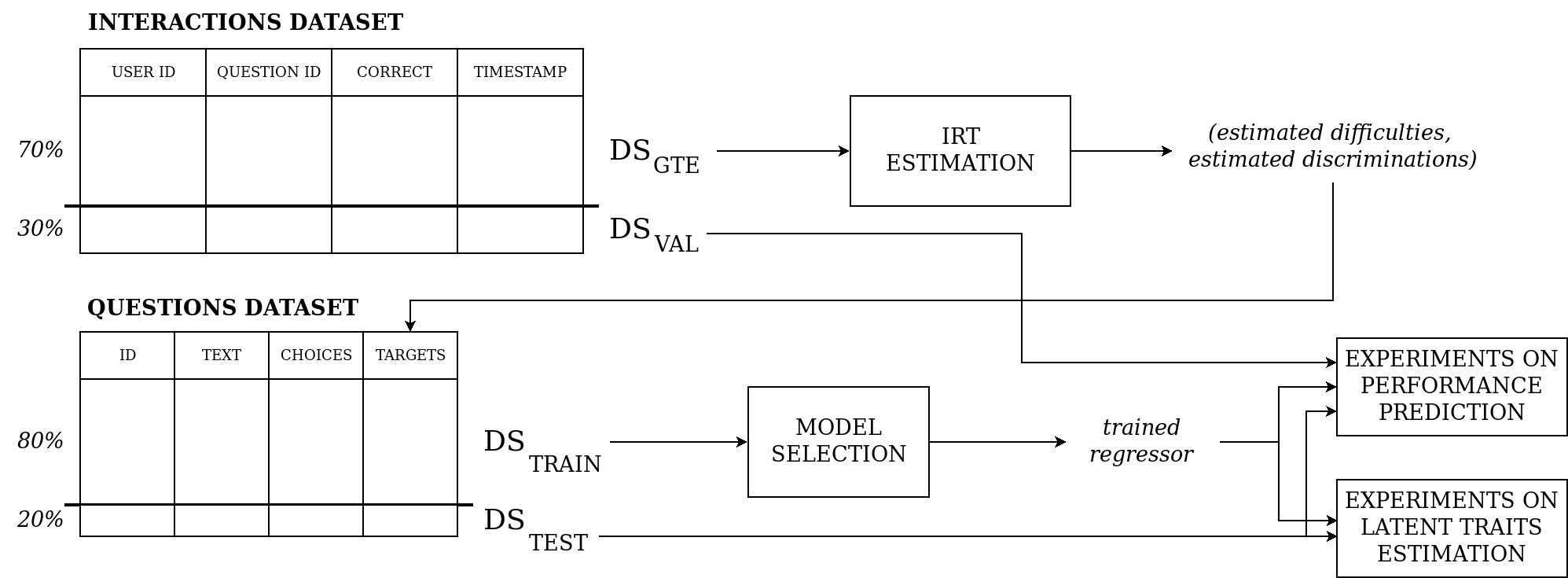}
\caption{Experimental setup. \label{fig:data_split}}
\Description[Experimental setup]{Experimental setup}
\end{figure*}

Training of our model is performed in two steps: i) an IRT model is trained in order to estimate the ``true'' difficulty and discrimination of each question, then ii) these IRT-estimated latent traits are used as target labels (\ie{} ground truth) to train \modelname{}, which gets the text as input. 
Therefore, in order to avoid any leaks of information between the training data and the test data, two different splits are performed on the dataset presented in Section \ref{dataset}.
Figure \ref{fig:data_split} displays how the four datasets are generated from the interaction dataset and the question dataset with the two splitting operations mentioned above.

First of all, a 70:30 split stratified on the questions is performed on the interaction dataset.
This leads to the generation of two smaller datasets: the interactions were randomly split, but we imposed the constraint of having at least one entry for each question; let us call the larger one $\text{DS}_{\text{GTE}}$, since it is used for the Ground Truth Estimation (\ie{} the initial estimation of the IRT latent traits), and the smaller one $\text{DS}_{\text{VAL}}$, since it will be used to validate the latent traits estimated with of \modelname{} on the performance prediction task.

The second split is performed on the questions of the question dataset with a 80:20 rate, thus generating two smaller datasets containing non-overlapping sets of questions.
The larger of the two question datasets generated from this split ($\text{DS}_{\text{TRAIN}}$) is used to train \modelname{}, while the smaller ($\text{DS}_{\text{TEST}}$) is used to test the estimation of the latent traits from the input text.

Thanks to this split performed in two different steps, it is possible to i) perform the ground truth estimation of the IRT latent traits using the interactions stored in $\text{DS}_{\text{GTE}}$; ii) train \modelname{} on $\text{DS}_{\text{TRAIN}}$;  iii) test its capability of estimating latent traits form text on $\text{DS}_{\text{TEST}}$; and iv) validate it on $\text{DS}_{\text{VAL}}$ by measuring its accuracy on the performance prediction task.

\section{Model Choice}\label{model_choice}
The ground truth latent traits of each question are estimated, using the interactions stored in $\text{DS}_{\text{GTE}}$, with a two-parameter IRT model implemented with \textit{pyirt}\footnote{\url{https://github.com/17zuoye/pyirt}}.
Then, several regression models are tested with five-fold cross validation in order to find the best configuration for each of them.
This is done using the input text of the questions in $\text{DS}_{\text{TRAIN}}$.
The following models are tested, and for all of them the scikit-learn\footnote{\url{https://scikit-learn.org/}} implementation is used:
\begin{itemize}
\item Random Forest (RF) Regressor \cite{breiman2001random}, with the following hyperparameters:
	\begin{itemize}
	\item \textit{n\_estimators} = [10, 25, 50, 100, 150, 200, 250]
	\item \textit{max\_depth} = [2, 5, 10, 15, 25, 50]
	\end{itemize}
\item Decision Tree (DT) Regressor \cite{breiman1984classification}, with the following hyperparameters:
	\begin{itemize}
	\item \textit{max\_features} = [1, 2, 3, 4, 5, None]
	\item \textit{max\_depth} = [2, 5, 10, 20, 50]
	\end{itemize}
\item Linear Regression (LR)\cite{seber2012linear}, with the following hyperparameters:
	\begin{itemize}
	\item \textit{normalize} = [True, False]
	\end{itemize}
\item Support Vector Regression (SVR)\cite{drucker1997support}, with the following hyperparameters:
	\begin{itemize}
	\item \textit{kernel} = ['linear', 'poly', 'rbf']
	\item \textit{gamma} = ['auto', 'scale']
	\item \textit{shrinking} = [True, False]
	\item \textit{degree} = [1, 2, 3, 4]
	\end{itemize}
\end{itemize}
Also, each configuration is tested on the three encodings presented in Subsection \ref{data_enc}: i) \encitem{}, ii) \enccorrect{}, iii) \encfull{}.
Lastly, each configuration is tested after performing dimensionality reduction with different values of \wt{}: in particular, values in the range $[100, 2000]$ are tested.
This preliminary experiment is performed twice, in order to choose the model configuration for both difficulty and discrimination estimation.

Table \ref{table:model_selection_diff} and Table \ref{table:model_selection_discr} display the results obtained with this initial experiment, showing the Mean Squared Error (MSE) for difficulty estimation and discrimination estimation respectively.
All the results presented in the table were obtained with the best performing configuration of hyperparameters and \wt{} for each ``family'' of models and for each encoding.

\begin{table}[ht]
\caption{Mean Squared Error for difficulty estimation, results of the preliminary experiment for model choice.}
\centering
\begin{tabular}{ c  c  c  c }
\hline
 & \encitem{} & \enccorrect{} & \encfull{} \\
Model & MSE & MSE  & MSE  \\
\hline
RF  & 0.359  & 0.306  & \textbf{0.306} \\
DT  & 0.790  & 0.767  & 0.757 \\
LR  & 0.719  & 0.691  & 0.669 \\
SVR  & 0.438  & 0.391  & 0.399 \\
\hline
\end{tabular}
\label{table:model_selection_diff}
\end{table}

\begin{table}[ht]
\caption{Mean Squared Error for discrimination estimation, results of the preliminary experiment for model choice.}
\centering
\begin{tabular}{ c  c  c  c }
\hline
 & \encitem{} & \enccorrect{} & \encfull{} \\
Model & MSE & MSE  & MSE  \\
\hline
RF  & 0.178  & 0.188  & \textbf{0.123} \\
DT  & 0.196  & 0.195  & 0.189 \\
LR  & 0.190  & 0.190  & 0.185 \\
SVR  & 0.187  & 0.188  & 0.200 \\
\hline
\end{tabular}
\label{table:model_selection_discr}
\end{table}

For both latent traits the best performing model was the Random Forest (RF) regressor, with the input text encoded using the \encitem{} encoding.
The configurations of hyperparameters leading to the lowest error, instead, were different for the two latent traits: i) for difficulty estimation, the RF is made of 250 estimators, each with a maximum depth of 50 and \wt{} is 1000; ii) for discrimination estimation the RF is composed of 100 estimators with 25 maximum depth; the considered \wt{} is 800.

\begin{figure}
\begin{subfigure}[t]{\linewidth}
  \centering
  \includegraphics[width=\linewidth]{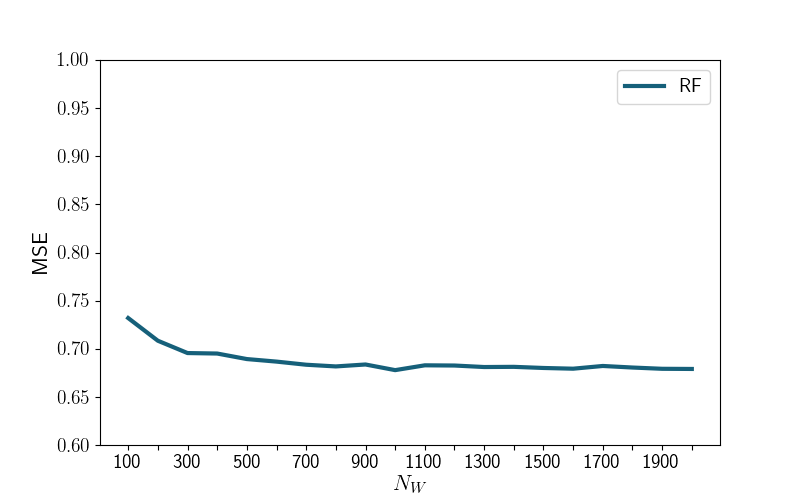}
  \caption{Mean Squared Error on difficulty estimation for varying \wt{}.}
  \label{fig:wt_effects_diff}
\end{subfigure}
\begin{subfigure}[b]{\linewidth}
  \centering
  \includegraphics[width=\linewidth]{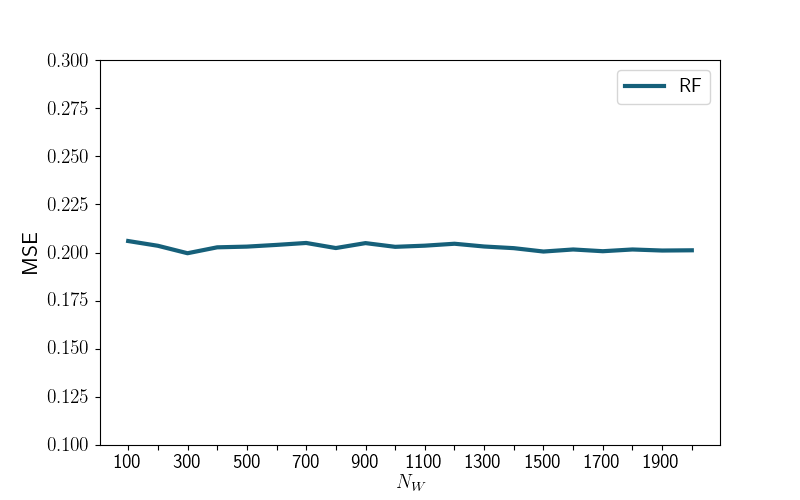}
  \caption{Mean Squared Error on discrimination estimation for varying \wt{}.}
  \label{fig:wt_effects_discr}
\end{subfigure}
\caption{Effects of varying \wt{} on the Mean Square Error while estimating difficulty and discrimination.}
\label{fig:wt_effects}
\Description[Effects of varying Nw]{Effect of varying Nw on the MSE}
\end{figure}

We also analyzed the effects of varying the value of \wt{} and observed that it does not have a significant impact on the error of the Random Forest.
Figure \ref{fig:wt_effects_diff} and Figure \ref{fig:wt_effects_discr} show the MSE on difficulty estimation and discrimination estimation respectively, obtained by the best configurations of the RF for varying value of \wt{}.

\section{Results}\label{results}
This section presents and discusses the results of the experiments carried on on the data.

\subsection{Latent Traits Estimation}\label{latent_traits_estimation}
First of all, we test the capability of the selected model configuration to estimate latent traits of new questions given the text of the question and the text of the possible answers.
This is done by estimating with \modelname{} the latent traits of the questions in $\text{DS}_{\text{TEST}}$ and comparing them with the IRT estimated values.
The Root Mean Squared Error (RMSE) and the Mean Absolute Error (MAE) for difficulty estimation and discrimination estimation are presented respectively in Table \ref{table:latent_traits_est_diff} and Table \ref{table:latent_traits_est_discr}.
The tables also show the relative errors (\ie{} Relative RMSE and Relative MAE), which represent the errors measured relatively to the range of possible values of difficulty and discrimination.
Specifically, the Relative RMSE for the difficulty is computed as:
\begin{equation*}
\text{Relative RMSE} = \frac{\text{RMSE}}{\text{max\_difficulty}-\text{min\_difficulty}}
\end{equation*}
Similarly, it can be computed the Relative RMSE for the discrimination and the Relative MAE for both latent traits.
For our experimental dataset, in particular, the IRT model was trained in order to have difficulties in the range $[-5; 5]$ and discriminations in the range $[-1; 2.5]$.

\begin{table}[ht]
\caption{Test of difficulty estimation.}
\centering
\begin{tabular}{ c c  c c }
\hline
 RMSE & Relative RMSE & MAE & Relative MAE \\
\hline
0.823 & 8.23\% & 0.639 & 6.39\%\\
\hline
\end{tabular}
\label{table:latent_traits_est_diff}
\end{table}

\begin{table}[ht]
\caption{Test of discrimination estimation.}
\centering
\begin{tabular}{ c c  c c }
\hline
 RMSE & Relative RMSE & MAE & Relative MAE \\
\hline
0.447 & 12.8\% & 0.329 & 9.4\%\\
\hline
\end{tabular}
\label{table:latent_traits_est_discr}
\end{table}

\begin{figure}[ht]
\centering
\includegraphics[width=\linewidth]{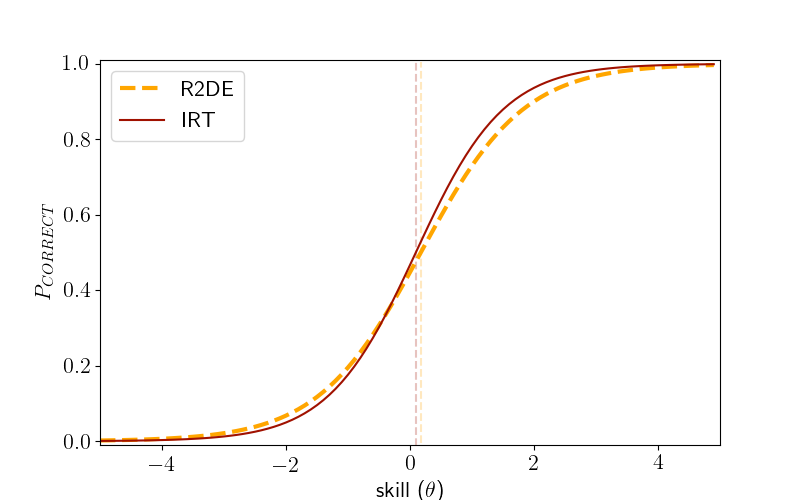}
\caption{Comparison between the item response function obtained with the latent traits estimated with IRT and with \modelname{}.}\label{fig:example_irf}
\Description[Comparison i.r.f.]{Comparison between the item response function obtained with the latent traits estimated with IRT and with R2DE}
\end{figure}

Figure \ref{fig:example_irf} shows, as example, the comparison between the item response function obtained with the IRT estimation and the one obtained with the latent traits estimated with \modelname{}.
The vertical lines represent the difficulty of the questions, implying the value estimated with IRT and the one estimated with \modelname{}.

Even though it is not possible to perform an actual comparison between the results of this work and previous ones, due to the fact that each research focused on a different (private) dataset, the analysis of the errors of the different approaches in difficulty estimation can still provide useful insight.
This cannot be done for discrimination estimation, since \modelname{} is the first model that is capable of estimating the discrimination as well as the difficulty of assessment items from the input text.
Table \ref{table:soa_comparison} compares the relative errors obtained in recent works.
The table shows that the error obtained in this work is smaller than the errors obtained in previous works and, although this does not assure that this model is better performing than the others on any datasets, it suggests that simple regression models as ours could perform as well as - maybe even better than - more complex models (\eg{} the Convolutional Neural Network with attention mechanism proposed in \cite{huang2017question}).

\begin{table}[ht]
\caption{Comparison with state of the art.}
\centering
\begin{tabular}{c c c c}
\hline
Paper & Difficulty range & RMSE & Relative RMSE\\
\hline
\modelname{} & [-5; 5] & 0.823 & \textbf{8.23\%} \\
Huang et al. \cite{huang2017question} & [0; 1] & 0.21 & \textbf{21\%} \\
Yaneva et al. \cite{yaneva2019predicting} & [0; 100] & 22.45 & \textbf{22.4\%} \\
\hline
\end{tabular}
\label{table:soa_comparison}
\end{table}

\subsection{Performance Prediction}
In Subsection \ref{latent_traits_estimation} we presented a comparison between the latent traits estimated with \modelname{} and the latent traits estimated with IRT, which are considered as ground truth.
However, this is not an observable ground truth and, for this reason, we also validate \modelname{} measuring its efficacy in the performance prediction task, which offers - as presented in Section \ref{background} - the only observable ground truth.
In particular, the interactions stored in $\text{DS}_{\text{VAL}}$ are used to validate the model.
The baselines that we use to test our model against are i) the ``ground truth'' latent traits estimated with IRT (which is a upper threshold), and ii) majority prediction.

Two different tests are carried out for performance prediction.
First, we filter the validation dataset ($\text{DS}_{\text{VAL}}$) in order to keep only the test questions (\ie{} the ones stored in $\text{DS}_{\text{TEST}}$).
Then, only the test questions are used both for measuring the accuracy of the prediction and for updating the estimation of the skill level at each step.
The results obtained with this experiment are presented in Table \ref{table:ppt_1} displaying for each approach the accuracy (Acc.), the precision (Prec.) and the recall (Rec.) on the correct interactions, and the precision and the recall on the wrong interactions.
The table shows that the latent traits estimated with \modelname{} lead, for most of the metrics, to values that are close to the ones obtained with ground truth latent traits and generally outperform majority prediction.
This means that \modelname{} is able to compute the latent traits from the question text with a performance similar to IRT that, instead, is based on hundreds of interactions.
\begin{table}[ht]
\caption{Results of the test on performance prediction, using only the questions in $\text{DS}_{\text{TEST}}$.}
\centering
\begin{tabular}{c c c c c c}
\hline
 & & \multicolumn{2}{l}{Correct Answers} & \multicolumn{2}{l}{Wrong Answers} \\
Approach & Acc. & Prec. & Rec. & Prec. & Rec. \\
\hline
\modelname{} & 0.662 & 0.704 & 0.794 & 0.562 & 0.442 \\ 
IRT          & 0.666 & 0.713 & 0.781 & 0.565 & 0.475 \\ 
Majority     & 0.625 & 0.625 & 1.0 & - & 0.0 \\ 
\hline
\end{tabular}
\label{table:ppt_1}
\end{table}

The second test explicitly reproduces the scenario that occurs in the wild.
In real assessment only some of the items are newly generated (thus requiring an estimation of the latent traits from text), most of them are already calibrated (\ie{} with known latent traits).
Therefore, in this second experiment on performance prediction both the test questions and the train questions of $\text{DS}_{\text{VAL}}$ are used.
However, only the test questions are considered for evaluating the performance of the model on performance prediction; the train questions are used exclusively to update the estimated skill level of the student during the experiment.
Anyway, the test questions are used also for updating the skill level estimation, as it was the case in the previous test.
Table \ref{table:ppt_2} displays the results obtained with this second experiment, using the same metrics as above.
Majority is not reported here since it is computed as in the previous test and performed much worse than the other two approaches.
Again, the latent traits estimated with \modelname{} proved good estimations for newly generated items.
Indeed, the accuracy obtained with our model is only 1.57\% lower than the accuracy obtained with IRT-estimated latent traits, which is the upper threshold.
\begin{table}[ht]
\caption{Results of the test on performance prediction on test interactions, skill estimated using all the questions.}
\centering
\begin{tabular}{c c c c c c}
\hline
 & & \multicolumn{2}{l}{Correct Answers} & \multicolumn{2}{l}{Wrong Answers} \\
Approach & Acc. & Prec. & Rec. & Prec. & Rec. \\
\hline
\modelname{} & 0.689 & 0.744 & 0.767 & 0.590 & 0.559 \\ 
IRT          & 0.701 & 0.757 & 0.768 & 0.603 & 0.589 \\
\hline
\end{tabular}
\label{table:ppt_2}
\end{table}

\section{Conclusions}\label{conclusions}
In this work, we introduced \modelname{}, a model which is capable of estimating the latent traits (\ie{} the difficulty and the discrimination) of newly generated multiple-choice questions by looking at the text of the question and the text of the possible choices.

Extensive experiments carried out on a large scale real world dataset provided by \ca{} showed that the model is capable of estimating with a low uncertainty both the difficulty and the discrimination.
Specifically, it reached a MAE of 0.639 for difficulty estimation (6.39\% of the whole difficulty range) and a MAE of 0.329 for discrimination estimation (9.4\% of the overall discrimination range).
\modelname{} is the first model estimating the discrimination as well as the difficulty, thus a comparison with the errors of other models was not possible for discrimination estimation.
On the other hand, a comparison with recent literature was performed for difficulty estimation; such comparison suggests that this model might be capable of performing at least as well as previously existing models.
However, an extensive comparison on the same datasets was not possible due to the unavailability of code from previous research and all the datasets being private.

We showed that this model improves the accuracy on the task of exam results prediction with respect to using a simple estimation such as majority estimation and reaches an accuracy comparable to the upper threshold obtained with the IRT-estimated latent traits.
Therefore, it is fair to say that \modelname{} reduces the importance of pretesting newly generated questions.
Indeed, the estimated latent traits only require some fine-tuning, which is much faster than performing pretesting from scratch involving few hundreds or few thousands students.
Moreover, having an estimation of the difficulty and the discrimination of the items is also useful for content creators at the time of writing the questions: they can have immediate feedback and modify the question accordingly, before deploying it.
Thus, this model enables a reduction in the number of questions that have to be removed from the set of assessment items due either to a too low discriminative power or to a too high (or too low) difficulty.

Future work will continue to explore this research direction, focusing in particular on the following aspects: i) using advanced embeddings for encoding the text of the assessment items; ii) analyzing the importance of specific keywords, in order to understand whether the very structure of questions and items - and not few keywords - is the reason why some questions are more difficult and more discriminative than others; iii) exploring the possibilities of explaining the predictions of the model.
Also, we aim at making possible a comprehensive comparison between this model and all the other models that use NLP approaches to estimate the latent traits of questions; for this reason, we make the code available for future research\footnote{\linkcode{}}, so that it might be run and tested on other datasets.


\bibliographystyle{ACM-Reference-Format}
\bibliography{bib}


\end{document}